\documentclass[letterpaper]{article} 
\usepackage{aaai25}  
\usepackage{times}  
\usepackage{helvet}  
\usepackage{courier}  
\usepackage[hyphens]{url}  
\usepackage{graphicx} 
\usepackage{natbib}  
\usepackage{caption} 
\usepackage{algorithm}
\usepackage{algorithmic}

\usepackage{newfloat}
\usepackage{listings}
\DeclareCaptionStyle{ruled}{labelfont=normalfont,labelsep=colon,strut=off} 
\floatstyle{ruled}
\newfloat{listing}{tb}{lst}{}
\floatname{listing}{Listing}
\usepackage{amsmath}
\usepackage{amssymb}
\usepackage{mathtools}
\usepackage{amsthm}
\allowdisplaybreaks[4]
\usepackage{cleveref}
\usepackage{booktabs}
\usepackage{multirow}
\usepackage{makecell}

\usepackage{color} 

\title{Learn How to Query from Unlabeled Data Streams in Federated Learning}
\author {
    Yuchang Sun\textsuperscript{\rm 1},
    Xinran Li\textsuperscript{\rm 1},
    Tao Lin\textsuperscript{\rm 2},
    Jun Zhang\textsuperscript{\rm 1}
}
\affiliations {
    \textsuperscript{\rm 1}Department of Electronic and Computer Engineering, The Hong Kong University of Science and Technology\\
    \textsuperscript{\rm 2}Westlake University\\
    \{yuchang.sun, xinran.li\}@connect.ust.hk,
    lintao@westlake.edu.cn, eejzhang@ust.hk
}

\begin{document}

\maketitle

\begin{abstract}
Federated learning (FL) enables collaborative learning among decentralized clients while safeguarding the privacy of their local data. Existing studies on FL typically assume offline labeled data available at each client when the training starts. Nevertheless, the training data in practice often arrive at clients in a streaming fashion without ground-truth labels. Given the expensive annotation cost, it is critical to identify a subset of informative samples for labeling on clients. However, selecting samples locally while accommodating the global training objective presents a challenge unique to FL. In this work, we tackle this conundrum by framing the data querying process in FL as a collaborative decentralized decision-making problem and proposing an effective solution named LeaDQ, which leverages multi-agent reinforcement learning algorithms. In particular, under the implicit guidance from global information, LeaDQ effectively learns the local policies for distributed clients and steers them towards selecting samples that can enhance the global model's accuracy. Extensive simulations on image and text tasks show that LeaDQ advances the model performance in various FL scenarios, outperforming the benchmarking algorithms.

\end{abstract}

\section{Introduction}

Federated learning (FL)~\cite{fedavg} has emerged as a distributed paradigm for training machine learning (ML) models while preserving local data privacy. 
In FL, the model training process involves multiple clients, each possessing its own local training data with different distributions. 
These clients jointly optimize an ML model based on their local data and periodically upload the model updates to the server.
Afterwards, the server updates the global model by applying the aggregated updates and distributes the current model to the clients for next-iteration training. 
This approach has gained significant attention in practical applications such as healthcare~\cite{RiekeH0MRABGLMO20} and Internet of Things~\cite{ZhangGHZKA22}, due to its potential to address privacy concerns associated with centralized data storage and processing.

Most of existing FL studies assume that clients have a fixed pool of training data samples with ground-truth labels~\cite{li2023revisiting,YeXWXCW23,YangXJ23,ShiLZTB23}, which, however, is unrealistic in many applications.
In practice, the data samples usually arrive at the clients without any label~\cite{fl_unlabeled}.
For example, hospitals collect raw medical images in the course of routine care which are initially unannotated raw images.
As labeling these images requires highly experienced clinical staff to spend minutes per image, it is more feasible to select a subset of samples for label querying and subsequent model training.
In this case, designing an effective strategy to identify the samples that could provide the most benefit in model training becomes a critical problem.

Data querying strategies for non-FL settings have been widely studied in the \emph{active learning} (AL) literature \cite{alsurvey0,alsurvey1,alsurvey2}. However, there are limited studies towards adapting these strategies to FL settings due to the challenges brought by the inherently decentralized nature of FL.
The major challenge lies in the conflict between the collaborative goal of optimizing the global model and the limited access to local data on clients.  
To be specific, FL aims to optimize the global model over all the training data across clients while each client can only access its private training data.
Given this fact, it is difficult for clients to select the most critical data samples which promote the global training. 
One straightforward approach is prioritizing local data on each client, which, however, is a myopic strategy without considering the goal of global model training, ultimately leading to suboptimal data selections.
More recently, there are some attempts to explore the data querying problem in FL with fixed unlabeled dataset by designing local selection rules~\cite{logo,kafal}. Nevertheless, the performance of these methods is restricted since the data selections on clients tend to operate in a non-cooperative fashion without the guidance towards the global training goal.
To solve these problems, it is imperative to design a specific data querying strategy tailored for FL with decentralized data streams while accounting for the collaborative objective of global model training.

\begin{figure}[!t]
    \centering
    \includegraphics[width=\columnwidth]{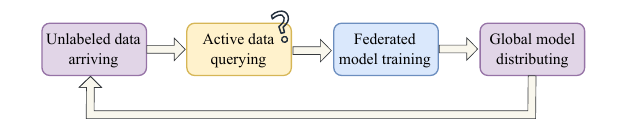}
    \caption{Workflow of the proposed LeaDQ framework.}
    \label{fig:workflow}
\end{figure}

In this work, we investigate the data querying problem in FL, particularly focusing on a case where unlabeled data arrive at clients in a streaming fashion. By alternating local data query and model training procedures, clients aim to collaboratively optimize the global model, as illustrated in Fig. \ref{fig:workflow}. In the subsequent sections, we first empirically show the impact of various data querying strategies on the performance of the global model. In particular, we find that a coordinated data querying strategy, which considers the training goal from a global perspective, yields greater performance benefit compared to individual data querying decisions. Motivated by this observation, we propose a data querying algorithm named \emph{LeaDQ} where clients query data in a decentralized manner while considering the global objective.
Specifically, we formulate the data querying problem in FL as a decentralized partially observable Markov decision process (Dec-POMDP) and leverage the multi-agent reinforcement learning (MARL) algorithms to learn local data querying policies for clients. By implicitly incorporating the status of the global model, the learned policies effectively guide the clients to make query decisions that are collaborative and ultimately beneficial for global training.
We conduct simulations to compare the LeaDQ algorithm with several representative data querying strategies. The experimental results on various image and text tasks demonstrate that LeaDQ selects samples that result in more meaningful model updates, leading to improved model accuracy.\footnote{The code is available at \url{https://github.com/hiyuchang/leadq/}.}

\section{Related Works}

\paragraph{Active learning}
AL algorithms are designed to select a subset of unlabeled data samples and query their labels from an oracle.
The design goal is to improve the model performance on the target dataset by selecting the most informative samples within the querying cost budget~\cite{alsurvey1,alsurvey2}.
The classic data selection metrics are uncertainty-based, diversity-based or the hybrid ones.
Specifically, the uncertainty-based AL algorithm~\cite{confidence_sampling} selects the unlabeled data with the lowest prediction confidence, which is straightforward for improving the performance of ML models. 
Meanwhile, the diversity-based method~\cite{coreset} aims to reduce redundancy in the selected instances for labeling, ensuring a more efficient use of labeling resources by avoiding similar samples.
Furthermore, the hybrid method attempts to combine the strengths of both uncertainty and diversity in the selection criteria.
In particular, some works~\cite{al_rl,batchal_marl} leverage reinforcement learning (RL) to learn a data query policy which is able to select the impactful data samples.

\paragraph{Federated active learning}
Recent studies have extended the investigations on data query problems to FL scenarios, termed as \emph{federated active learning} (FAL)~\cite{logo,kafal,FAL,FAL2,FAL3}.
In FAL, clients select some local data samples and ask the oracles to label these samples.
\citet{logo} design a two-step algorithm named \textbf{LoGo} to select informative data samples while resolving the inter-class diversity on clients.
Meanwhile, \textbf{KAFAL}~\cite{kafal} leverages the specialized knowledge by the local clients and the global model to deal with their objective mismatch.
Nevertheless, the clients only have access to their own data and tend to query the samples that are beneficial for their local training objective.
Because of the discrepancy between the local and global training objectives, the queried data may bring less benefit for the global model training.
Moreover, these works assume that all unlabeled data samples are available beforehand and require evaluating them at every time.
Different from these works, we focus on a more challenging stream-based setting, where the unlabeled samples sequentially arrive at clients and may be accessed only once.

In addition, another thread of works~\cite{stream_fl_async,stream_fl,GongZ0SLC23,Shi2023self} designs various approaches to deal with the unlabeled data or streaming data in FL.
These works, however, are beyond the scope of this paper, since they tackle other aspects such as self-supervised learning or data storage on clients instead of the data querying problem.

\section{Problem Formulation}\label{sec:problem_formulation}

We consider an FL system where $K$ clients with unlabeled data streams cooperate to train an ML model by alternating data querying and model training.
The goal of clients is to minimize the following objective:
\begin{equation}
    \min_{\theta\in\mathbb{R}^M} \mathbb{E}_{(\mathbf{x},y) \sim \mathcal{P}} [\ell(\theta; \mathbf{x},y)],
\end{equation}
where $\ell(\theta; \mathbf{x},y)$ denotes the loss value of model $\theta$ on data sample $(\mathbf{x},y)$ with data input $\mathbf{x}$ and corresponding label $y$. 
Note that the target data distribution $\mathcal{P}$ is unknown and inaccessible.
We assume that the aggregated training data on all clients follow the same distribution as that of the target dataset.
Nevertheless, the local data distribution $\mathcal{P}_k$ of each client $k\in[K]$ is typically different from the global data distribution, which is also known as the non-independent and identically distributed (non-IID) setup in FL~\cite{kairouz2021advances}.
Formally, the assumptions on data distributions are given as: $\cup_{k\in[K]} \mathcal{P}_k = \mathcal{P}$ and $\mathcal{P}_k \neq \mathcal{P}, \forall k \in [K]$.

\begin{figure*}[t]
    \centering
    \includegraphics[width=\linewidth]{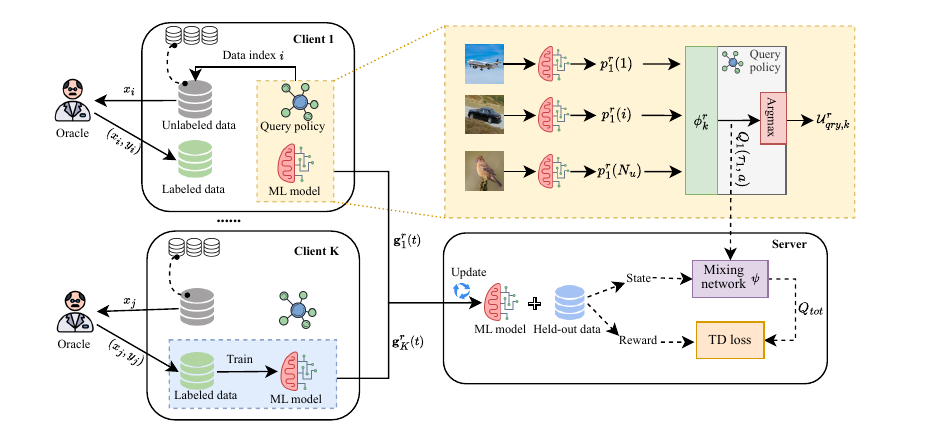}
    \caption{An overview of the proposed LeaDQ framework. The yellow block is the process of active data querying and the blue block is the process of federated model training. The dashed lines are conducted only when training the query policies.}
    \label{fig:fal}
\end{figure*}

Raw training data without ground-truth labels arrive at client $k$ according to the underlying distribution $\mathcal{P}_k$~\cite{stream_fl}.
According to the data arrival status, the training process of optimizing model $\theta$ can be divided into $R$ rounds.
At the beginning of each round $r\in\{1,2,\dots,R\}$, some new unlabeled data samples arrive at each client.
To utilize these data for model training, clients actively query the labels of some selected samples from an oracle.
Afterwards, they collaboratively train the ML model with the current labeled dataset.
The above process is illustrated in Fig.~\ref{fig:fal}.
In the following, we introduce the procedure of data querying and model training in detail.

\paragraph{Active data querying}
In the $r$-th round, a set of $N_u$ unlabeled data samples arrives at client $k$, denoted by $\mathcal{U}_k^r = \{\mathbf{x}_i\}_{i=1}^{N_u}$ where $\mathbf{x}_i$ is the feature of the $i$-th sample.
Each client then makes a binary decision $\mathbf{a}_k^r = \{0,1\}^{N_u}$, indicating whether a particular sample is selected for labeling.\footnote{In practical applications, the number of arrived data samples and queried samples may vary among clients or across training rounds. We assume these values are constant in this work for simplicity of expression but our method can be easily adapted to more general cases.}
Specifically, $\mathbf{a}_k^r[i]=1$ if the $i$-th sample is chosen for querying and $\mathbf{a}_k^r[i]=0$ otherwise.
If being queried, an oracle provides the ground-truth label $y_i \in\mathcal{Y} $ for this sample $\mathbf{x}_i$.
We denote the number of samples for labeling per round on each client as $N_q$ and the queried sample set on client $k$ as $\mathcal{U}_{qry,k}^r$. 
Afterwards, the client updates its local training set $\mathcal{L}_k^{r-1}$ by incorporating the newly labeled data samples, i.e.,
\begin{equation}
    \mathcal{L}_k^{r} = \mathcal{L}_k^{r-1} \cup \mathcal{U}_{qry,k}^r.
    \label{eq:labeled}
\end{equation}

Given the high labeling cost, only a subset of unlabeled samples is queried for annotation. The labeling budget of each client is expressed as:
\begin{equation}
    |\mathcal{U}_{qry,k}^r| = N_q.
\end{equation}
The data querying policy, parameterized with a \emph{query model} $\phi$, outputs the selection decisions $\mathbf{a}_k^r$ for the given unlabeled dataset $\mathcal{U}_k^r$ on client $k$.
It is worth noting that the query model may vary among clients and across rounds depending on the training stage and the current status of ML model.

\paragraph{Federated model training}
Based on the local labeled data $\mathcal{L}_k^r$, clients train the current global model $\theta^r$ to optimize the following objective:
\begin{equation}
    \min_{\theta^r\in\mathbb{R}^M} f(\theta^r) \triangleq \sum_{k=1}^{K} \frac{|\mathcal{L}_k^r|}{|\mathcal{L}^{r}|} F_k(\theta^r; \mathcal{L}_k^r),
\end{equation}
with the local loss function
\begin{equation}
    F_k(\theta^r; \mathcal{L}_k^r) \triangleq \frac{1}{|\mathcal{L}_k^{r}|} \sum_{(\mathbf{x}, y) \in \mathcal{L}_k^{r}} \ell(\theta^r;\mathbf{x}, y).
\end{equation}
Here $\mathcal{L}^{r} = \bigcup_{k=1}^{K} \mathcal{L}_{k}^{r}$ denotes the labeled dataset on all clients.
As the training data contain privacy-sensitive information and cannot be exposed to the server, we adopt the classic FedAvg algorithm~\cite{fedavg} to optimize the ML model. 
Specifically, the model training process in one round can be divided into $T$ training iterations.
At the beginning, the clients initialize the current model as\footnote{Instead of using the trained model from previous round, some works also reset the model to random or designed weights for next-round training~\cite{warm_start}.} $\theta^r(0) = \theta^{r-1}$. In the $t$-th training iteration, client $k$ performs local training for multiple steps to compute the accumulated gradients $\mathbf{g}_k^r(t)$.
The gradients are then uploaded to the server for aggregation. The global model is then updated using these gradients as:
\begin{equation}
    \theta^r(t+1) = \theta^r(t) - \eta \sum_{k=1}^{K} \mathbf{g}_k^r(t), \forall t=0,\dots,T-1.
    \label{eq:global_update}
\end{equation}
After $T$ training iterations, the global model is finalized as $\theta^{r} = \theta^r(T)$.

Through iterative data querying and model training, the ML model is optimized by clients.
However, the conflict remains between the clients' shortsighted view on its local data and the target data distribution of the global model.
The queried data may be therefore redundant or not that beneficial for the global model training. 
To address this problem, it is intuitively helpful to provide the clients with a global view that captures the current training status and implies the desired data for facilitating the global model training. 
To clearly demonstrate this phenomenon, we present an example in the next section.

\section{Motivating Example}\label{sec:motivation}
In this section, we show empirically that a data querying strategy with a global view can benefit the FL training. In particular, we explore an FL setup in which clients collaborate to classify data samples from the MNIST dataset by training a LeNet model~\cite{mnist}.
We assume there are $N_u=10$ unlabeled data samples arriving at each client in each round and only one data sample is selected for labeling.
The classical coreset method~\cite{coreset} selects the samples such that the model learned over the selected samples is competitive for the remaining data points.
For an unlabeled dataset $\mathcal{U}$ and given labeled dataset $\mathcal{L}$, the data sample to be queried is selected as: 
\begin{equation}
    \mathbf{x}^* =  \arg\max\limits_{\mathbf{x} \in \mathcal{U}} \min_{\mathbf{x}^\prime \in\mathcal{L}} \left\| \mathbf{x} - \mathbf{x}^\prime\right\|_2.
    \label{eq:coreset}
\end{equation}

We first implement the coreset approach on each client locally, i.e., $\mathcal{U}=\mathcal{U}_k^r$, termed as \textbf{Local Coreset}.
In Local Coreset, clients make selections based on the local available data samples.
Besides, we consider an ideal approach termed as \textbf{Global Coreset} where the server can access to all clients' data $\mathcal{U}=\sum_{k\in[K]} \mathcal{U}_k^r$ and determine one queried sample for each client.
After querying the labels, clients iteratively train the ML model based on the labeled dataset.
The experimental details are referred to Appendix B.1.

\begin{figure}[!t]
    \centering
    \includegraphics[width=\columnwidth]{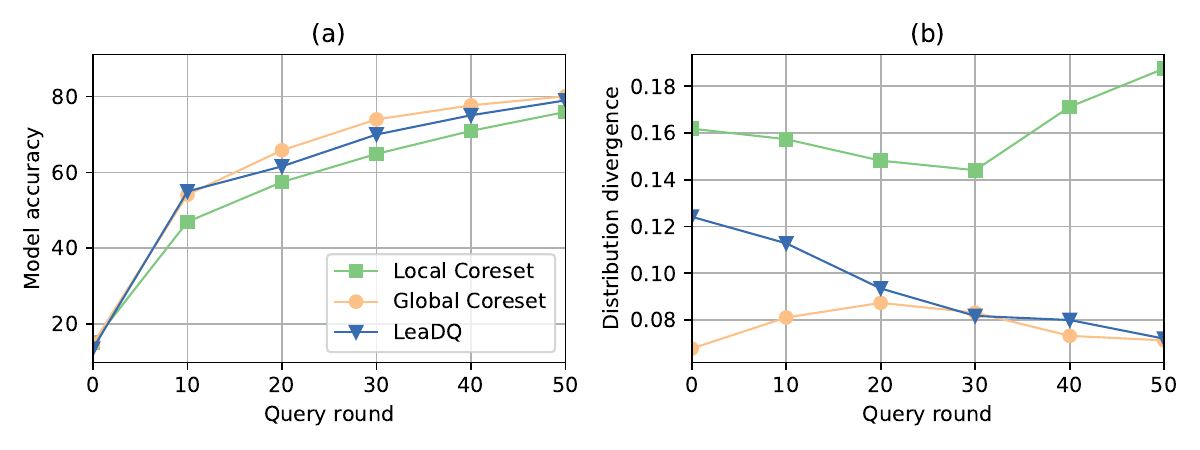}
    \caption{Comparisons between different data querying strategies. In (a), the model accuracy is computed on the test data; in (b), the \textit{distribution divergence} is computed as the Kullback–Leibler (KL) divergence between the distributions of selected unlabeled data on all clients and the target data.}
    \label{fig:motivation}
\end{figure}

The evaluation in Fig.~\ref{fig:motivation} focuses on the effectiveness of various data querying strategies and the distribution of queried data samples.
It is observed that the Global Coreset strategy improves the model accuracy compared with Local Coreset, benefiting from the utilization of global information when selecting unlabeled data.
This can be verified by the distribution divergence between the queried samples in each round and the global data distribution as shown in Fig.~\ref{fig:motivation} (b).
In particular, Global Coreset can select critical samples that coincide with the target data distribution, since it leverages the information from a global view.

Nevertheless, the global information in Global Coreset requires computing the sample-wise distance between each unlabeled sample and labeled sample.
As the data samples are distributed on different clients, it is impractical to directly implement the Global Coreset, which severely violates the data privacy of clients.
Thus, it becomes crucial to devise a strategy that leverages global insights without necessitating direct access to data across all clients.
To this end, we propose the LeaDQ algorithm under a centralized training decentralized execution (CTDE) paradigm, which implicitly incorporates the global objective into the querying strategies while maintaining the decentralized execution as in FL. Consequently, LeaDQ achieves comparable performance (as shown in Fig.~\ref{fig:motivation}) to Global Corset while preserving the data privacy.

\section{LeaDQ: Learning the Data Querying Policies for Clients}\label{sec:method}
Effective data querying strategies in FL should exhibit the ability to select unlabeled samples that are beneficial for global model training, while making the decisions in a decentralized fashion.
Motivated by the capabilities of MARL algorithms in making individual decisions while optimizing a collaborative objective~\cite{vinyals2017starcraft,wang2021multi}, we propose an MARL-based data querying algorithm named \textbf{LeaDQ}, short for \emph{\underline{Lea}rn \underline{D}ata \underline{Q}uerying}. 
In the following, we first formulate the data querying problem in FL as a Markov decision process and elaborate on the details of system design in Section~\ref{sec:decision}.
Subsequently, we detail the process of decentralized data querying and centralized policy training in Sections~\ref{sec:exe} and~\ref{sec:training}, respectively.

\subsection{Data Querying in FL as A Decentralized Decision-Making Process}\label{sec:decision}
The objective of data querying per client $k$ in FL aims to select samples from local unlabeled dataset $\mathcal{U}_k^r$ to query their labels in round $r$.
By data querying and model training, these clients aim to collaboratively optimize the global model $\theta^r$.
However, each client only has access to its local data without the knowledge of either data distribution of other clients or the target global data distribution, i.e., the decision making process is based on partial observation in its nature.

To describe such a setup, we frame the data querying problem as a decentralized partially observable Markov decision process (Dec-POMDP) \cite{dec-POMDP} denoted by a tuple $\langle \mathcal{S}, \mathcal{A}, P, R, \mathcal{O}, K, \gamma\rangle$.
We view each querying round as a discrete timestep and each client as an agent in this process.

At each discrete timestep $r$, a \textbf{global state} $\mathbf{s}^r \in \mathcal{S}$ reveals the current status of global model training.
We assume the server owns a held-out dataset $\mathcal{D}_{\text{held}}$~\cite{public1,public2} and the global state is defined as the \emph{prediction confidence} of the global model $\theta^r$ on it:
\begin{equation}
    \mathbf{s}^r = \left[\max_{y\in\mathcal{Y}} p(y| \mathbf{x}; \theta^r) \right]_{\mathbf{x} \in \mathcal{D}_{\text{held}}} \,.
    \label{eq:state}
\end{equation}
We assume the held-out dataset follows the same distribution as that of the whole dataset, and thus the global state defined in \eqref{eq:state} serves as a low-cost indicator to show the current model's performance on the target dataset and can effectively reveal the current training status.

When a set of unlabeled data $\mathcal{U}_k^r$ with the length of $N_u$ arrives, each client $k$ gets the \textbf{local observation} $\mathbf{o}_k^r \in \mathcal{O}$ of these data.
Inspired by previous works \cite{confidence_sampling}, we use the \textit{predictive logits} output by the current model $\theta^r$ as the local observation, i.e.,
\begin{equation}
     \mathbf{o}_k^r = \left[ l(\mathbf{x}; \theta^r)\right]_{\mathbf{x} \in \mathcal{U}_k^r},
     \label{eq:local_obs}
\end{equation}
where $l(\mathbf{x}; \theta^r)= p(y| \mathbf{x}; \theta^r), \forall y\in\mathcal{Y}$ encapsulates the predicted logits.
The local observation $\mathbf{o}_k^r$ is a low-dimension representation of arrived samples in $\mathcal{U}_k^r$ and also reflects the uncertainty of samples. It is intuitive that samples with higher levels of uncertainty would be more helpful to model updates when being included in the training dataset.

Based on its local observation $\mathbf{o}_k^r$, each client chooses \textbf{actions} $\mathbf{a}_k^r = \{0,1\}^{N_u}$ to select samples for labeling. Here the binary decision $\mathbf{a}_k^r[i]=1$ denotes the case of the $i$-th sample being chosen for querying and $\mathbf{a}_k^r[i]=0$ otherwise.
The joint action is then given by $\mathbf{A}^r = [\mathbf{a}_1^r, \mathbf{a}_2^r,\dots,\mathbf{a}_K^r]$.
After querying unlabeled data, the clients cooperatively update the ML model based on the labeled dataset following the procedure elaborated in Section~\ref{sec:problem_formulation}.
Such evolution of the ML model's states can be characterized by the transition function $P(\mathbf{s}' | \mathbf{s}, \mathbf{a}) : \mathcal{S} \times \mathcal{A} \rightarrow [0, 1]$.

Afterwards, the updated global model returns to clients a joint \textbf{reward} $R^r$, which reveals the impact of selected actions on the ML model training.
As the collaborative goal of clients is to improve the overall performance of the ML model $\theta^r$, this reward should reveal its current training status.
Thus we define the reward as the difference of model accuracy on the held-out data before and after updating, i.e.,
\begin{equation}
    R^r \equiv R(\mathbf{s}^{r}, \mathbf{A}^{r}) = Acc(\theta^{r}; \mathcal{D}_{\text{held}}) - Acc(\theta^{r-1}; \mathcal{D}_{\text{held}}).
    \label{eq:reward}
\end{equation}
The reward in \eqref{eq:reward} shapes a \emph{global} view of the model training, which helps clients to make better \emph{local} decisions of data querying.
Formally, clients aim to find the optimal data querying policies that maximize the total discounted reward $R_{tot}^{r} \!=\! \sum_{j=r}^{r+J} \gamma^j R^j$ with episode length $J$ and discount $\gamma$.

By trial-and-error, the data query policies are directed to incorporate the information of the global model training, indicated by state $\mathbf{s}^r$ and reward $R^r$, into the collaborative decisions, ultimately promoting the global model training.
When executing the query policies, it is sufficient for clients to select local unlabeled data samples in a decentralized manner.
In summary, such CTDE strategy bridges the gap between the local data querying and the objective of global model training, as detailed in the following subsections.

\begin{algorithm}[!t]
\caption{The LeaDQ framework}
\label{alg:proposed}
\begin{algorithmic}[1]
\FOR{each round $r = 1$ to $R$} 
    \STATE \textit{// Active data querying}
    \FOR{each client $k = 1$ to $K$}
        \STATE Compute predictive logits as local observations $\mathbf{o}_k^r$ \hfill \COMMENT{$\triangleright$~Eq.~\ref{eq:local_obs}}
        \STATE Compute local Q-values as $Q_k(\mathbf{a},\mathbf{o}_k^r; \phi_k^r)$
        \STATE Choose actions $\mathbf{a}_k^r$ according to greedy policy \hfill \COMMENT{$\triangleright$~Eq.~\ref{eq:choose_action}}
        \STATE Query the oracle for label $y_i$ of data sample $\mathbf{x}_i$ with $a_k^r[i] = 1$
        \STATE Update the labeled dataset as $\mathcal{L}_k^{r}$ \hfill \COMMENT{$\triangleright$~Eq.~\ref{eq:labeled}}
    \ENDFOR
    
    \STATE \textit{//Federated model training}
    \STATE Initialize the global model as $\theta^{r}(0) = \theta^{r-1}$
    \FOR{each iteration $t = 0$ to $T-1$}
        \FOR{each client $k = 1$ to $K$}
            \STATE Train the current model $\theta^{r}(t)$ based on the local labeled dataset $\mathcal{L}_k^{r}$
            \STATE Upload the gradients $\mathbf{g}_k^r(t)$ to the server
        \ENDFOR
        \STATE The server updates the global model as $\theta^r(t+1)$ \hfill \COMMENT{$\triangleright$ Eq.~\ref{eq:global_update}}
    \ENDFOR
    \STATE Update the global model as $\theta^{r} = \theta^r(T)$
\ENDFOR
\end{algorithmic}
\end{algorithm}

\subsection{Decentralized Policy Execution for Querying Data Samples}\label{sec:exe}
In round $r$, client $k$ has access to its local action-observation history $\tau_k \in \mathcal{T} \equiv (\mathcal{O} \times \mathcal{A})^*$.
Based on such history, the client makes a decision of selecting unlabeled data following its data query policy.
Specifically, each client feeds the local observation $\mathbf{o}_k^r$ of arrived unlabeled data into its policy network, denoted by $\phi_k^r$, to output the local Q-value.
Subsequently, it chooses the action that induces the highest local Q-value, which is given by:
\begin{equation}
    \mathbf{a}_k^r = \arg\max_{\mathbf{a}:|\mathbf{a}|=N_q} Q_k(\mathbf{a}, \mathbf{o}_k^r; \phi_k^r),
    \label{eq:choose_action}
\end{equation}
where $|\mathbf{a}|=N_q$ implies $N_q$ unlabeled samples are queried.
Depending on the value of $\mathbf{a}_k^r$, each unlabeled sample is determined for querying or not.
As such, the clients use the query policy to select data in a decentralized manner without violating the data privacy.
The details of the proposed framework are summarized in Algorithm~\ref{alg:proposed}.

When querying data, the clients rely on their local policies, which shall guide them to select critical samples for the global model training.
To integrate the global information into this process, we adapt the QMIX algorithm~\cite{qmix} into the LeaDQ framework for training policies.

\begin{figure*}[!t]
    \centering
    \includegraphics[width=\textwidth]{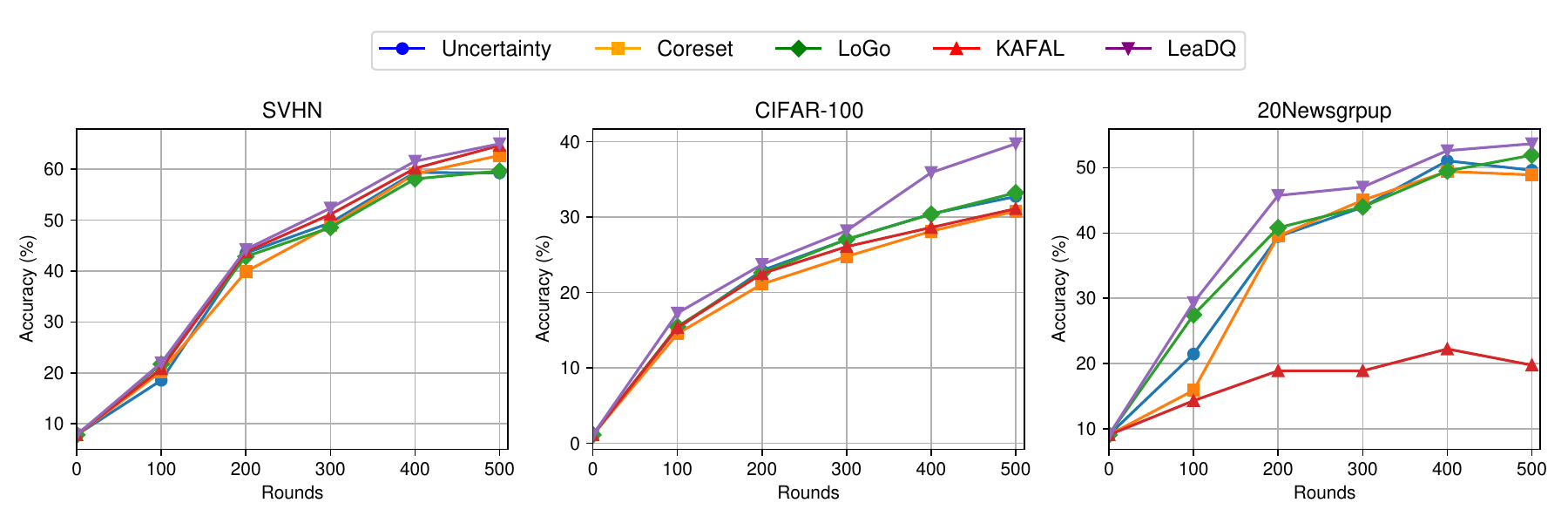}
    \caption{Model accuracy (\%) in different data querying rounds.}
    \label{fig:main}
\end{figure*}

\subsection{Centralized Policy Training for Learning Local Policies}\label{sec:training}
By selecting unlabeled data, all clients contribute to the global model training and get a shared reward, serving as the global feedback for the joint decision of clients.
However, it is difficult for clients to directly use the shared reward for local data selections.
To link the global feedback and local decisions, we define the joint Q-value as a weighted sum of the local Q-values, which is given by~\cite{qmix}:
\begin{equation}
    Q_{tot} = \psi(Q_1, \cdots, Q_K, \textbf{s}^r). 
    \label{eq:joint-q}
\end{equation}
Here $\psi$ is a \emph{mixing network} from a family of monotonic functions to model the complex interactions between collaborating clients.
This mixing network helps to coordinate the clients' local policies while accounting for the state of current ML model $\textbf{s}^r$.

We use a replay buffer $\mathcal{B}$ to store the transition tuple $\langle \mathbf{s}, \mathbf{o}, \mathbf{a}, R, \mathbf{s}', \mathbf{o}' \rangle$, where the state $\mathbf{s}'$ is observed after taking the action $\mathbf{a}$ in state $\mathbf{s}$ and receiving reward $R$.
During policy training, we iteratively sample batches of transitions from the replay buffer and update the policy network by minimizing the temporal difference (TD) loss:
\begin{equation}
    \begin{split}
    & \mathcal{L}(\{\phi_k\}, \psi) \\
    & = \mathbb{E}_{(\mathbf{s}, \mathbf{o}, \mathbf{a}, R, \mathbf{s}', \mathbf{o}' ) \sim \mathcal{B}} \left[ \left( y^{tot} - Q_{tot}(\mathbf{s}, \mathbf{o}, \mathbf{a}; \{\phi_k\}, \psi) \right)^2\right],
    \end{split}
    \label{eq:td-loss}
\end{equation}
where $y^{tot} = R + \gamma \max_{\mathbf{u}} Q_{tot}(\mathbf{s}', \mathbf{o}', \mathbf{u}; \{\phi_k\}, \psi)$ with discount $\gamma$.
Indeed, minimizing the TD loss in \eqref{eq:td-loss} optimizes the local policies $\{\phi_k\}$ on clients towards a global goal, i.e., increasing the reward $R$. As such, by following their learned policies, clients can select the beneficial unlabeled data for global model training.
The pseudo code of policy training is deferred to Appendix A.1.

\section{Experiments}\label{sec:experiments}
\subsection{Setup}\label{sec:exp_setup}

We simulate an FL system with one server and $K=10$ clients.
We evaluate the algorithms on two image classification tasks, i.e., SVHN~\cite{svhn} and CIFAR-100~\cite{cifar10}, and one text classification task, i.e., 20Newsgroup~\cite{20news}.
We train a convolutional neural network (CNN) model with four convolutional layers~\cite{logo} on the SVHN dataset, a ResNet-18~\cite{resnet} on the CIFAR-100 dataset, and a DistilBERT~\cite{distilbert} model on the 20Newsgroup dataset.
To simulate the non-IID setting, we allocate the training data to clients according to the Dirichlet distribution with concentration parameter $\alpha=0.5$~\cite{silos}.
In each round, $N_u=10$ unlabeled data samples arrive at each client independently and each client selects $N_q=1$ data sample for label querying.
The details of datasets are summarized in Table~\ref{tab:datasets}.
More implementation details can be found in Appendix B.2. Extended results can be found in Appendix C.
In all tables, the best performances are highlighted in \textbf{bold}, and the second-best ones are \underline{underlined}.

\begin{table}[h]
\centering
\caption{Summary of datasets.}
\label{tab:datasets}
\renewcommand{\arraystretch}{1.2}
\begin{tabular}{c|c|c}
\toprule
Type & Dataset & Non-IID Type \\ \midrule
Text & 20Newsgroup & Distribution-based \\ \midrule
\multirow{2}{*}{Image}  & SVHN & \makecell[c]{Distribution-based \\ Quantity-based} \\
\cline{2-3}
 & CIFAR-100 & Distribution-based \\
\bottomrule
\end{tabular}
\end{table}

\paragraph{Baselines}
We compare the proposed algorithm with the following baselines.
First, we adapt several representative AL methods to FL, including: (i) \textbf{Uncertainty}~\cite{confidence_sampling}: Each client selects the samples with lowest prediction confidence for label querying; (ii) \textbf{Coreset}~\cite{coreset}: Each client selects the samples such that the model learned over these samples is competitive for the remaining unlabeled data points.
Second, two state-of-the-art federated AL algorithms are also compared, including: (iii) \textbf{LoGo}~\cite{logo}: Each client selects samples using a clustering-based sampling strategy; (iv) \textbf{KAFAL}~\cite{kafal}: Each client selects samples with the highest discrepancy between the specialized knowledge by the local model on clients and the global model.

\subsection{Main Results}

We show the model accuracy in different querying rounds on several datasets in Fig. \ref{fig:main}. We observe that the proposed LeaDQ algorithm achieves the best accuracy in all the querying rounds.
On the SVHN dataset, the naive local querying strategies (i.e., Uncertainty and Coreset) attain weaker performance than other approaches, showcasing the importance of designing specific strategies for FL.
On the CIFAR-100 dataset, the baselines have similar performance, while LeaDQ surpasses the best-performance baseline at the later training stage by around 6\%, implying its advantages in querying critical samples for global model training.

\begin{figure}[!t]
    \centering
    \includegraphics[width=0.48\textwidth]{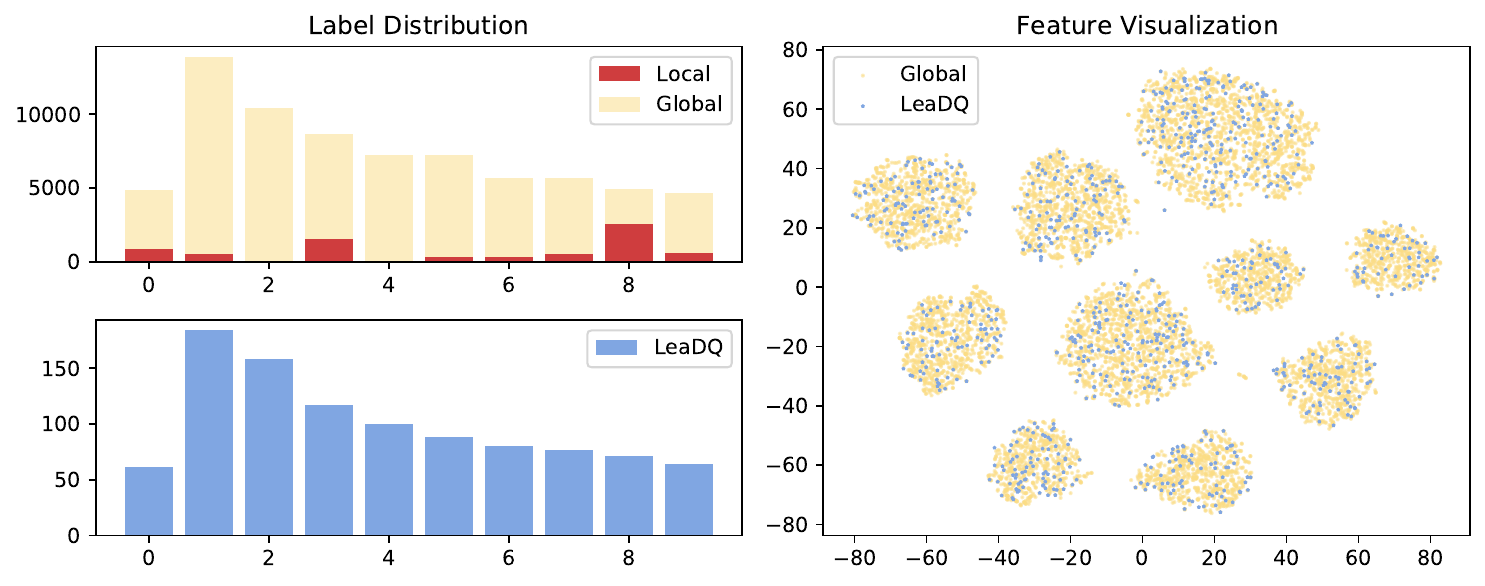}
    \caption{Visualization of the data samples in SVHN dataset ($R=100$). (a) Label distribution of local data on a random client, target global data, and queried data by LeaDQ; (b) T-SNE plots~\cite{tsne} of feature distributions of the global data and queried data by LeaDQ.}
    \label{fig:visualization}
\end{figure}

\paragraph{Visualization of queried data samples}
To visualize the querying strategies provided by LeaDQ, we show the queried samples distributions in Fig.~\ref{fig:visualization}.
According to Fig.~\ref{fig:visualization} (a), the queried unlabeled data samples have similar distributions with the target global data dsitribution, despite the fact that the local data follows a different distribution with the global data. 
Besides, the t-SNE plots in Fig.~\ref{fig:visualization} (b) also demonstrate that LeaDQ is able to select unlabeled data that have a similar feature distribution with the target global data.

\subsection{Results With Diverse Scenarios}

\paragraph{Results with various arrived and queried data samples}
We show the model accuracy with different numbers of arrived samples $N_u$ and queried samples $N_q$ per round in Table~\ref{table:query-ratio}.
We observe from the results that as the ratio of queried data samples $N_q/N_u$ increases, the model achieves higher accuracy with all the data querying strategies.
This is because leveraging more labeled data samples on clients effectively improves the model's performance.
In most cases, LeaDQ outperforms the baselines due to its advantages in identifying critical samples.
Meanwhile, when the number of queried samples is larger ($N_q >1$), LeaDQ and KAFAL have competitive performances, surpassing other strategies.

\begin{table}[!t]
\centering
\caption{Model accuracy (\%) on SVHN with different values of arrived samples $N_u$ and queried samples $N_q$.}
\label{table:query-ratio}
\begin{tabular}{lccccc}
\toprule
$N_q / N_u$ & $1/30$ & $1/20$ & $1/10$ & $2/10$ & $3/10$ \\ \midrule
Uncertainty & 44.35      & \underline{50.64}      & 59.28  & 73.31 & 79.30  \\
Coreset     & 39.96      & 45.07      & 62.72 & 75.97 & 79.66 \\
\midrule
LoGo        & \underline{45.45}      & 49.94      & 59.66 & 76.02 &  79.86     \\
KAFAL       & 41.88      & 49.30      & \underline{64.67}  &   \textbf{76.32}    &  \underline{80.30}     \\
\midrule
LeaDQ       & \textbf{45.68}      & \textbf{52.31}      & \textbf{65.02} &   \underline{76.28}    &  \textbf{80.37} \\
\bottomrule
\end{tabular}
\end{table}

\begin{table}[!t]
\centering
\caption{Model accuracy (\%) on SVHN with different degrees of non-IID under distribution-based skew. $\alpha$ is the value of concentration parameter in Dirichlet distribution, 
and a larger $\alpha$ indicates that the data distribution across clients is closer to IID.}
\label{table:dirichlet}
\begin{tabular}{lccc}
\toprule
& $\alpha=0.1$ & $\alpha=0.5$ & $\alpha=1.0$ \\ \midrule
Uncertainty & 48.09 & 59.28 & 63.54 \\
Coreset & 49.77 & 62.72 & 65.04 \\
\midrule
LoGo & 43.70 & 59.66 & 59.97 \\
KAFAL & \underline{52.14} & \underline{64.67} & \underline{66.48} \\ \midrule
LeaDQ & \textbf{53.98} & \textbf{65.02} & \textbf{66.50} \\
\bottomrule
\end{tabular}
\end{table}

\begin{table}[!t]
\centering
\caption{Model accuracy (\%) on SVHN with different degrees of non-IID under quantity skew. $C$ is the number of classes on each client, and a larger $C$ indicates that the data distributions across clients are closer to IID.}
\label{table:quantity}
\begin{tabular}{lccc}
\toprule
& $C=2$ & $C=3$ & $C=4$ \\ \midrule
Uncertainty & 62.26 & 67.73 & 67.10  \\
Coreset & \underline{64.15} & \underline{67.77} & \underline{68.43} \\
\midrule
LoGo & 59.69 & 63.15 & 67.07 \\
KAFAL & 20.19 & 20.32 & 21.75 \\ \midrule
LeaDQ & \textbf{65.34} & \textbf{68.19} & \textbf{69.06} \\
\bottomrule
\end{tabular}
\end{table}

\paragraph{Results with different setups of data heterogeneity}
In addition, we simulate two types of data heterogeneity, i.e., the distribution-based skew and quantity skew~\cite{silos}.
The results are illustrated in Tables~\ref{table:dirichlet} and~\ref{table:quantity}, respectively, in which we show the model accuracy after $R=500$ rounds when applying different data querying strategies.
We vary the data heterogeneity among clients by adjusting the concentration parameter $\alpha$ and the number of classes on each client $C$.
When the data tend to be more heterogeneous, the model performance degrades as the training becomes harder.
We find that LeaDQ achieves higher performance compared with the baselines in all cases, showing its effectiveness in finding beneficial samples for global model training.
However, the FAL baselines may even fail in identifying informative samples and thus lead to suboptimal performance. Overall, these results demonstrate the robustness and adaptability of the proposed LeaDQ algorithm. 

\section{Conclusions}

In this paper, we investigate the data querying problem in stream-based federated learning, highlighting the conflict between local data access and global querying objective. We introduce a novel MARL-based algorithm named LeaDQ, which learns local data query policies on distributed clients. LeaDQ promotes cooperative objectives among clients, ultimately leading to improved machine learning model training. Extensive experiments validate the superiority of LeaDQ over state-of-the-art baselines, demonstrating its effectiveness in querying critical data samples and enhancing the quality of the ML model.
We note that LeaDQ provides a framework for data querying in FL.
Future research could explore the incorporation of more advanced algorithms beyond FedAvg and QMIX to enhance performance further. \looseness=-1

\section*{Acknowledgements}
This work was supported by the Hong Kong Research Grants Council under the Areas of Excellence scheme grant AoE/E-601/22-R and NSFC/RGC Collaborative Research Scheme grant CRS\_HKUST603/22.

\bibliography{aaai25}

\clearpage
\appendix
\onecolumn
\section{Additional Details for Implementing LeaDQ}

\subsection{Pseudo Codes for Policy Training}\label{appendix:pseudo}

We show the pseudo codes for policy training in LeaDQ in~\cref{alg:policy}.

\begin{algorithm}[h]
\caption{Policy training in LeaDQ}
\label{alg:policy}
\begin{algorithmic}[1]
\FOR{each round $r = 1$ to $R$}
    \STATE Conduct \textit{Active data querying} to get local observations $\mathbf{o}_k^r$, joint action $\mathbf{A}^r$ \hfill \COMMENT{$\triangleright$~Algorithm~1}
    \STATE Conduct \textit{Federated model training} to get new model $\theta^{r}$ \hfill \COMMENT{$\triangleright$ Algorithm 1} 
    \STATE The server computes the global state $\mathbf{s}^{r}$ and reward $R(\mathbf{s}^{r}, \mathbf{A}^{r})$ based on held-out data $\mathcal{D}_{held}$
    \STATE The server stores the transition tuple $\langle \mathbf{\tau}, \mathbf{s}, \mathbf{a}, R, \mathbf{s}' \rangle$ in the replay buffer $\mathcal{B}$
    \IF{the number of stored transitions is larger than the batch size}
        \FOR{update step $i= 1$ to maximal\_update\_steps}
            \STATE Compute the TD loss \hfill \COMMENT{$\triangleright$~Eq.~13}
            \STATE Update the mixing network $\psi$ and the local policy network $\phi_k^r$
        \ENDFOR
    \ENDIF
\ENDFOR
\end{algorithmic}
\end{algorithm}

\subsection{A General Case for Querying Multiple Data Samples}\label{appendix:general}
Each client selects a subset of $N_q$ unlabeled data samples, denoted by $\mathcal{U}_{\text{qry}}$, to be labeled. In the LeaDQ framework, the querying strategy is adapted so that the clients take an action $\mathbf{a}_k^r$ where the size of the action, denoted by $|\mathbf{a}_k^r|$, equals $N_q$. This action is aimed at maximizing the aggregate of local Q-values, which is given by:
\begin{equation}
    \mathbf{a}_k^r = \arg\max_{|\mathbf{a}|=N_q} Q_k(\mathbf{a}, \mathbf{o}_k^r; \phi_k^r).
    \label{eq:choose_action_sum}
\end{equation}
The summation in the above equation ensures that the action selected corresponds to the highest total Q-values. Consequently, this results in identifying the most beneficial data samples for labeling according to the given policy.

\section{Experimental Details}\label{appendix:details}
We implement all methods with PyTorch and run all experiments on NVIDIA RTX 3080Ti GPUs.

\subsection{Implementation Details for the Motivating Example}\label{appendix:motivation}
We simulate the FL system with $30$ clients and evaluate the algorithms on the MNIST dataset with a LeNet with two convolutional layers~(LeCun et al. 1998.
The training data are allocated to clients according to the Dirichlet distribution with concentration parameter $0.5$~(Li et al. 2022).
The methods are implemented as follows:
\begin{itemize}
    \item \textbf{Local Coreset}: Each client selects unlabeled data locally according to~Eq.~7.
    \item \textbf{Global Coreset}: The server collects all unlabeled data samples from all clients and selects some samples for label querying.
    It initially selects one sample for each client and iteratively finds samples via a greedy furthest-first traversal conditioned on all labeled examples.
\end{itemize}

\paragraph{Metrics}
The \textit{distribution divergence} is compute as the Kullback–Leibler (KL) divergence between the distributions of selected unlabeled data on all clients (denoted by $\mathcal{U}_{\text{qry}}^r$) and the target data (denoted by $\mathcal{U}_{tgt}$), which is given by:
\begin{equation}
    Div(\mathcal{U}_{\text{qry}}^r; \mathcal{U}_{tgt}) = \sum_{y\in\mathcal{Y}} P_{\text{qry}}(y) \log\left(\frac{P_{\text{qry}}(y)}{Q_{tgt}(y)}\right),
\end{equation}
where $P_{\text{qry}}(y)$ is the label distribution of data in $\mathcal{U}_{\text{qry}}^r$, and $Q_{tgt}(y)$ is the label distribution of data in $\mathcal{U}_{tgt}$.


Other implementations are same as that of the main experiments, as explained in the next subsection.

\subsection{Implementation Details for Main Experiments}
\paragraph{Training task}
We evaluate the algorithms on two image classification tasks, i.e., SVHN~(Netzer et al. 2011) and CIFAR-100 (Krizhevsky and Hinton 2009), and one text classification task, i.e., 20Newsgroup (Lang 1995), each of which contains a training set and a test set.
In the default setup, the training set is allocated to clients according to the Dirichlet distribution with concentration parameter $0.5$.
The ML model is evaluated on the test set as the performance metric.
Main experiment results are averaged over three random seeds.

\paragraph{Details of federated model training}
We use FedAvg~(McMahan et al. 2017) as the local training algorithms on clients.
For fair comparisons, we adopt the same hyper-parameters for all methods, as shown in~\cref{table:fedavg}.

\begin{table*}[!t]
    \centering
    \caption{Implementation details.}
    \label{table:fedavg}
    \begin{tabular}{lccc}
        \toprule
         Dataset & SVHN & CIFAR-100 & SVHN \\ \midrule
         Model & CNN & ResNet-18 & DistilBERT  \\ 
         Batch size & 64 & 64 & 64  \\
         Learning rate & 0.01 & 0.01 & 0.01 \\
         Local epochs  & 1    & 1 & 1 \\
         Initial labeled samples & 10 & 100 & 10 \\
         Arrived samples per round   & 10    & 10 & 10 \\
         Training iterations & 30 & 30 & 10 \\
         Model reset & Random weights & Last-round weights & Last-round weights \\
         Total querying rounds & 500 & 500 & 500 \\
         Total arrived samples in Table 1 & 50000 & N/A & N/A \\
         \bottomrule
    \end{tabular}
\end{table*}

\paragraph{Implementation of LeaDQ}
We use Recurrent neural network (RNN) as agent network and use Double-Q update strategy.
We summarize the main hyper-parameters in LeaDQ algorithm in~\cref{table:marl}.
In the warm-up stage, LeaDQ selects the local samples according to the prediction scores of samples, which is given by:
\begin{equation}
    \mathbf{x}^*
    = \arg\min_{\mathcal{U}_{\text{qry}}\subset \mathcal{U} } \max_{y\in\mathcal{Y}} \sum_{\mathbf{x}\in\mathcal{U}_{\text{qry}}} p(y|\mathbf{x}; \theta).
\end{equation}

\begin{table}[ht]
\caption{Implementation details of LeaDQ.}
\label{table:marl}
\centering
\begin{tabular}{ll}
\toprule
\textbf{Hyperparameter}          & \textbf{Value}                \\ \midrule
Discount $\gamma$ & 0.99 \\
Episode length $J$  & 10 \\
Replay buffer size               & 1000 episodes                 \\
Warm-up timesteps                     & 32                         \\
Batch size                       & 32                   \\
Optimizer                        & Adam                       \\
Learning rate                    & 0.01          \\
Held-out dataset size & 1000 \\
\midrule
Double-Q update & True \\
Parameter sharing among clients & True \\
Q-Network update frequency & 1 \\
Q-Network maximal update steps & 200 \\
\bottomrule
\end{tabular}
\end{table}

\paragraph{Metrics}
The \emph{model accuracy} is evaluated on the test dataset in different data querying rounds. 
The \emph{T-SNE} (short for t-Distributed Stochastic Neighbor Embedding) plots visualize the distribution of feature representations extracted by a fine-tuned ResNet-18 model, in a 2D space~(Van der Maaten and Hinton 20).

\section{Supplementary Results}

\subsection{Results on the Tiny-ImageNet Dataset}
Table~\ref{table:imagenet} shows the model accuracy after 200 querying rounds on the Tiny-ImageNet dataset. The results further verify the benefits and robustness of LeaDQ in achieving better model accuracy.

\begin{table}[h]
\centering
\caption{Model accuracy (\%) on the Tiny-ImageNet Dataset}
\label{table:imagenet}
\begin{tabular}{lcc}
\hline
             & $\alpha=0.5$ & $\alpha=0.1$ \\
\hline
Uncertainty & 20.32          & 20.20          \\
Coreset     & \underline{21.52}          & 20.48          \\
LoGo        & 21.50          & \underline{21.05}          \\
KAFAL       & 16.70          & 16.88          \\
LeaDQ       & \textbf{21.68} & \textbf{22.00} \\
\hline
\end{tabular}
\end{table}

\subsection{Results with More Clients.}
We compare the model performance when there are 50 clients in the system. The results in Table~\ref{table:50clients} show that LeaDQ consistently leads to the best model accuracy, showing its scalability.

\begin{table}[h]
\centering
\caption{Model accuracy (\%) with 50 clients on the SVHN dataset.}
\resizebox{0.5\textwidth}{!}{
\label{table:50clients}
\begin{tabular}{lccccc}
\hline
 Setup & Uncertainty & Coreset & LoGo & KAFAL & LeaDQ \\
\hline
50 clients & 73.42 & \underline{73.57} & 71.37 & 44.77 & \textbf{74.09} \\
\hline
\end{tabular}}
\end{table}

\subsection{Hyper-parameter Analysis}
We study the effect of episode length $J$ on the performance of our proposed LeaDQ method.
The following results on the SVHN dataset in Table~\ref{tab:J} show that episode length has minimal impact on the final accuracy, highlighting the robustness of LeaDQ to this parameter.

\begin{table}[h]
\centering
\caption{Model accuracy for different values of $J$. }
\label{tab:J}
\begin{tabular}{lccc}
\hline
J & 5 & 20 & 30 \\
\hline
Acc. & 61.00\% & 59.79\% & 59.79\% \\
\hline
\end{tabular}
\end{table}

\section{Discussions}\label{appendix:limitation}

\paragraph{Discussions on privacy in LeaDQ}
The core idea of FL is to train an ML model from distributed data on clients. (a) By sharing models instead of data, FL preserves the local data privacy of clients. (b) Meanwhile, additional techniques such as secure aggregation further protect the privacy of each client's model.
In this work, we study the data querying problem in FL. To preserve the same privacy as in FL, the data cannot be accessed by other clients or the server. Thus, it is hard for clients to select proper samples for labeling based on solely local data. We propose an MARL-based algorithm named LeaDQ to solve this problem. In a nutshell, LeaDQ allows clients to query data in a decentralized manner while maintaining the privacy protection in FL.

\end{document}